\documentclass[letterpaper, 10 pt, conference]{ieeeconf}
\IEEEoverridecommandlockouts
\usepackage{amsmath}
\usepackage{amssymb}
\usepackage{amsfonts}
\usepackage{graphicx}
\usepackage[normalem]{ulem}
\usepackage{xcolor}

\title{\LARGE \bf
Dynamic Watermarking and Action-Conditioned Latent Dynamics for Secure Vision-Based Controllers
}

\author{Aditya Parameshwaran$^{1}$, Navid Aftabi$^{2}$, Abhishek Hanchate$^{3}$, \\ Satish T. S. Bukkapatnam$^{3}$, Dan Li $^{2}$ 
and Yue Wang$^{1*}$%
\thanks{$^*$ Corresponding Author}
\thanks{$^{1}$Department of Mechanical Engineering, Clemson University, Clemson, SC 29634 , USA. {\tt\small aparame@clemson.edu, yue6@clemson.edu}}%
\thanks{$^{2}$Department of Industrial and Systems
Engineering, University of Washington, Seattle, WA 98195 USA {\tt\small aftabi@uw.edu; dli27@uw.edu}}%
\thanks{$^{3}$ Wm
Michael Barnes ’64 Department of Industrial and Systems Engineering, Texas A\&M University, College Station, TX 77843 USA {\tt\small abhishek.hanchate@tamu.edu; satish@tamu.edu}}%
}

\usepackage{amsthm}
\theoremstyle{plain} 
\newtheorem{theorem}{Theorem}

\newtheorem{lemma}{Lemma}

\begin{document}

\maketitle
\thispagestyle{empty}
\pagestyle{empty}

\begin{abstract}
Replay attacks are difficult to detect in vision-based control systems because previously recorded observations can remain plausible while the physical system follows a different trajectory. This paper develops an adaptive dynamic-watermarking framework for detecting replay attacks against vision-based controllers. A private Gaussian watermark is added to the nominal control input, and a control input-conditioned latent dynamics network (LDN) predicts the corresponding evolution in a compact latent space learned from image observations. Attack evidence is obtained from the discrepancy between the predicted latent response and the latent representation of the received image observation. A reinforcement learning policy adjusts the watermark covariance using the latent state and detector confidence, thereby balancing attack detectability, watermark energy, and control distortion. A Lagrangian penalty on task-specific safety costs is included during training to limit the adverse effect of the excitation. The framework is evaluated on a vision-guided UR5 manipulation task in which the robot reaches a target while avoiding a moving human arm under nominal and replay-attacked observations.
\end{abstract}

\section{Introduction}
Vision-based neural controllers are powerful in cluttered, human-robot settings, but their sensing channel is vulnerable to replay attacks: an adversary can replace the current image stream with previously recorded observations that remain visually plausible. In safety-critical tasks, such attacks can conceal a physical deviation while the robot continues to act on stale visual feedback. Dynamic watermarking (DWM) addresses this by injecting a private excitation into the control input and checking whether the current measurements contain the corresponding response~\cite{mo2009secure,satchidanandan2017dynamic}. Existing methods typically form this consistency check from known state-space dynamics and low-dimensional measurements~\cite{porter2021generalized,naha2023sequential}. That assumption does not directly apply to a black-box neural controller where the observations can be high-dimensional images with unknown dynamics. We address this gap with a control input-conditioned latent dynamics network (LDN) that predicts the evolution of image representations in a compact visual latent space, enabling watermark-response residuals to be formed without requiring an explicit analytical model of the image dynamics ~\cite{parameshwaran2025sevin,wu2023daydreamer}. 

\begin{figure*}[ht]
    \centering
    \vspace*{6pt}
    \includegraphics[width=0.95\textwidth]{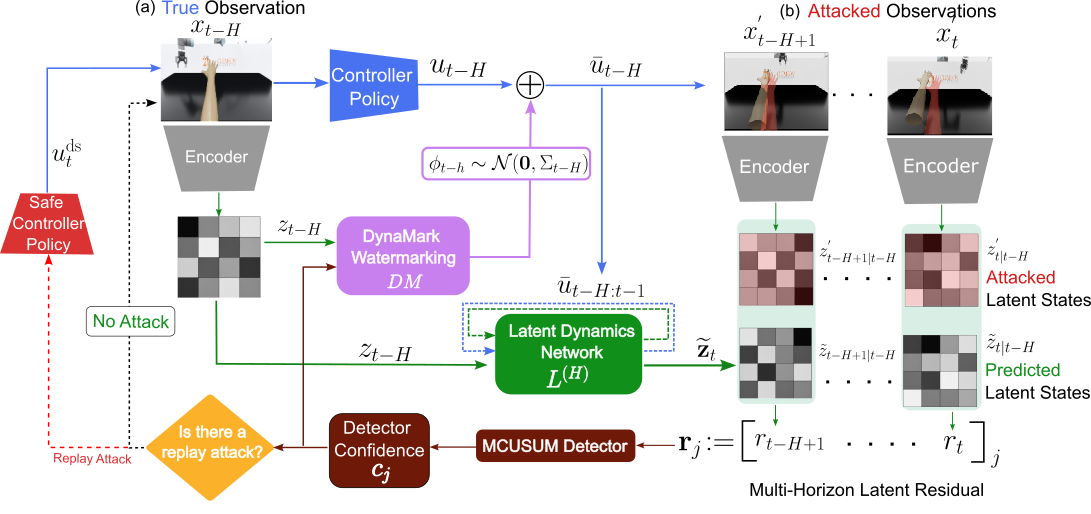}
    \caption{Overview of the proposed framework. (a) The live observation $x_{t-H}$ is provided to the vision-based controller policy $F$, which predicts the control input $u_{t-H}$. In parallel, an encoder maps $x_{t-H}$ to the latent state $z_{t-H}$. The latent state, together with the detector confidence $c_j$, is provided to the DWM policy to select the covariance $\Sigma_{t-H}$, from which the private watermark $\phi_{t-H}$ is sampled. The watermark and control input are combined to form the watermarked control input $\bar{u}_{t-H}$, which is applied to the plant. (b) When a replay attack occurs, the encoder maps the received replayed observations to latent states $[z'_{t-H+1},\ldots,z'_t]$. These are compared with the predicted latent states $[\tilde z_{t-H+1|t-H},\ldots,\tilde z_{t|t-H}]$ to obtain the residual window $\mathbf{r}_j$, defined in~\eqref{eq:latent_residual}. The MCUSUM detector processes this window, and the resulting evidence updates the attack confidence $c_j$ through~\eqref{eq:ct}. When attack evidence persists, the supervisor switches to the damped-stop controller and applies $u_t^{\mathrm{ds}}$ as specified in~\eqref{eq:supervisory_command}.}
    \label{fig:approach}
\end{figure*}

We condition the LDN on a privately watermarked control input and compare its predicted latent response with the latent representation of the received observation. The resulting latent space residual provides replay evidence, while a learned DWM policy adapts the watermark covariance to balance detectability, actuation effort, and task performance. The framework thereby combines latent dynamics prediction, adaptive watermarking, and supervisory fallback in a single vision-based replay-defense architecture as seen in Figure~\ref{fig:approach}.

Our contributions are summarized as follows:
\begin{enumerate}
    \item We detect replay attacks by verifying a private DWM signal through predictions from a control input-conditioned LDN. 
    \item We characterize how replay-induced watermark mismatch alters the MCUSUM detector statistic and develop a Bayesian belief update that quantifies online detection confidence and serves as a compact input to the DWM policy.
    \item We train the DWM policy with a Lagrangian penalty on robot--human separation violations and evaluate supervisory stopping under replay. We demonstrate the framework in a simulated human-robot task where misleading visual feedback can lead to unsafe motion. 

\end{enumerate}

\section{Related Work}
\subsection{Replay Attacks and Dynamic Watermarking}
Replay attacks are particularly challenging because prerecorded measurements can remain plausible after the physical system has diverged, concealing unsafe behavior. Similar threats arise in visual verification systems, where replayed images can mislead face age verification~\cite{korshunov2024replay}, and detection methods combine spatial and temporal features to identify video replay attacks~\cite{khan2025spatiotemporal}. Dynamic watermarking (DWM) helps detect replay attacks by injecting a private random excitation and checking whether the measurements reflect the system’s response to the current control input ~\cite{mo2009secure}. Prerecorded measurements that remain plausible will fail this response test and thus reveal the attack's occurrence. This principle was then formalized with innovation-based detector statistics for linear time-invariant and time-varying systems~\cite{satchidanandan2017dynamic,porter2021generalized}, and later extended to sequential detection rules that balance information gain and control cost~\cite{naha2023sequential}. These methods provide the theoretical foundation for replay detection, but their residuals are generally derived from explicit plant and observation models with low-dimensional measurements. For a black-box neural controller operating directly on images, neither an analytical transition model nor a tractable image-space measurement model is readily available. Our approach replaces that requirement with a learned control input-conditioned latent dynamics model, allowing the private watermark response to be tested directly from visual observations.

The excitation must also be selected without unnecessarily degrading control performance. Naha \emph{et al.} choose a stationary watermark variance by trading off information gain and control cost, while DynaMark learns a time-varying Gaussian policy using the operating condition and Bayesian detector confidence~\cite{aftabi2025dynamark}. Our work retains this detector-dependent adaptation but evaluates dynamical consistency using learned image representations instead of low-dimensional state estimates. We further add a task-specific safety cost and a supervisory fallback to an observation-independent controller when the attack confidence exceeds a threshold.

\subsection{Latent Dynamics Models for Visual Control}

In visual control, latent dynamics models provide a compact representation in which action-conditioned dynamics can be rolled forward over multiple steps. PlaNet uses learned latent dynamics for online planning from pixels~\cite{hafner2019planet}, while Dreamer learns behavior through imagined latent trajectories~\cite{hafner2020dreamer}; DayDreamer demonstrates the use of world models for learning directly on physical robots~\cite{wu2023daydreamer}. TD-MPC further shows how task-oriented latent dynamics can support short-horizon trajectory optimization for continuous control~\cite{hansen2022tdmpc}. Building on this line of work, our LDN autoregressively predicts future image-derived latents, feeding each predicted latent and the next privately watermarked control input into the subsequent prediction. We compare this predicted trajectory with latents encoded from subsequently received images; their discrepancy supplies replay evidence to the detector. The detector confidence and current latent state then inform adaptive watermark selection.

\section{Methodology}

\subsection{Problem Setup}
Consider an image-based neural network controller $F:\mathcal{X}\times\mathbb{R}^{d_p}\to\mathbb{R}^{d_u}$, where $x_t\in\mathcal{X}\subseteq\mathbb{R}^{d_x}$ is the image observation and $x_t^{\mathrm{prop}}\in\mathbb{R}^{d_p}$ is a trusted proprioceptive measurement. The nominal control command is
\begin{equation}
    u_t = F(x_t,x_t^{\mathrm{prop}}), \qquad u_t\in\mathbb{R}^{d_u}.
\end{equation}
An encoder $E:\mathcal{X}\to\mathbb{R}^{d_z}$ maps each received image to a compact latent representation:
\begin{equation}
    z_t = E(x_t), \qquad z_t \in \mathbb{R}^{d_z},
\end{equation}
where $d_z\ll d_x$. The dimensions $d_x$, $d_p$, $d_z$, and $d_u$ denote the image, proprioceptive, latent, and control dimensions, respectively.

Physical watermarking injects a private random signal $\phi_t$ into the nominal control input, producing
\begin{equation}\label{eq:u_phi}
    \bar u_t = u_t + \phi_t.
\end{equation}
We treat $\phi_t$ as a private signal sampled online from a Gaussian distribution whose covariance $\Sigma_t$ is selected dynamically at runtime. While existing approaches commonly employ a fixed static watermark distribution~\cite{mo2009secure,satchidanandan2017dynamic,naha2023sequential}, we use an adaptive dynamic-watermarking strategy inspired by DynaMark~\cite{aftabi2025dynamark}. The DynaMark policy and its RL formulation are described in Section~\ref{subsec:dynamark_rl}.  Under honest sensing, the observation received after applying $\bar u_t$ contains the physical response to the current private realization $\phi_t$. A replayed observation was generated under a different, earlier control input and watermark realization, so it generally lacks the response associated with the current $\phi_t$. The detector tests this mismatch by predicting the latent response to the control input that was actually applied and comparing that prediction with the latent encoding of the subsequently received image. At each step, the watermark is sampled as $\phi_t=\Sigma_t^{1/2}\xi_t$, where the draws $\xi_t\sim\mathcal{N}(\mathbf{0},I_{d_u})$ are independent and $\Sigma_t$ is selected before drawing $\xi_t$. The historical watermark $\phi_t'$ associated with a replayed image was generated using an earlier draw and may have a different covariance.

Let the latent dynamics network be
\begin{equation}\label{eq:ldn_map}
    L:\mathbb{R}^{d_z}\times\mathbb{R}^{d_u}\to\mathbb{R}^{d_z},
\end{equation}
where $L(z_t,\bar u_t)$ predicts the next latent state $z_{t+1|t}$ from the current latent state and watermarked control input applied during $t$ and $t+1$. Over a prediction horizon $H$, let $L^{(H)}$ denote the recursive rollout of the same LDN under the applied control input sequence.
Specifically, the rollout is initialized with $\tilde{z}_{t-H|t-H}=z_{t-H}$ and satisfies $\tilde{z}_{t-H+h|t-H} = L\left(\tilde{z}_{t-H+h-1|t-H}, \bar{u}_{t-H+h-1}\right)$ for $h=1,\dots, H$. The corresponding stacked latent prediction is

\begin{align}
    \widetilde{\mathbf z}_t
    :&=\mathrm{vec}\left(\tilde z_{t-H+1|t-H},\ldots,\tilde z_{t|t-H}\right) \notag\\
    &=L^{(H)}\!\left(z_{t-H},\bar u_{t-H:t-1}\right).\label{eq:ldn}
\end{align}

Here, $\tilde z_{t-H+h|t-H}$ is the latent prediction for discrete instant $t-H+h$, corresponding to physical time $(t-H+h)\Delta t$, made from the latent state available at instant $t-H$.
At the window-completion time \(t\), the detector has received \(z_t\) and has stored the applied control input sequence \(\bar{u}_{t-H:t-1}\). It can therefore complete the rollout initialized at \(t-H\).

To protect the robot when attack evidence persists, we use a Simplex-style supervisor that switches from the watermarked RL controller to a trusted damped-stop controller. Let $c_t$ denote the detector confidence, let $\nu_t$ count consecutive detector updates for which $c_t\ge c_{\mathrm{sw}}$ (and reset to zero otherwise), and let $N_{\mathrm{p}}$ be the required persistence length. The supervisor mode is $\gamma_t:=\mathbb{I}\{\nu_t\ge N_{\mathrm{p}}\}$. At the switching instant $t_s$, the fallback initializes its reference velocity from the trusted end-effector velocity measurement, $v_{t_s}^{\mathrm{ds}}=v_{t_s}^{\mathrm{ee}}$, and then applies the discrete damping law
\begin{equation}\label{eq:damped_stop_velocity}
    v_{t+1}^{\mathrm{ds}}
    =
    \left(I-\Delta t K_d\right)v_t^{\mathrm{ds}},
    \qquad K_d\succ0.
\end{equation}
We choose $K_d$ so that $\rho(I-\Delta t K_d)<1$, where $\rho$ denotes the spectral radius. This makes the reference velocity converge to zero. Because the nominal controller interface accepts target poses, the fallback converts the damped velocity reference into a compatible end-effector position target,
$u_t^{\mathrm{ds}}=p_t^{\mathrm{ee}}+\Delta t\,v_t^{\mathrm{ds}}$, while holding the end-effector orientation fixed. The control input sent to the low-level controller is therefore
\begin{equation}\label{eq:supervisory_command}
    u_t^{\mathrm{cmd}} =
    \begin{cases}
        \bar u_t, & \gamma_t=0,\\
        u_t^{\mathrm{ds}}, & \gamma_t=1.
    \end{cases}
\end{equation}
Thus, after detection the supervisor bypasses the RL target-pose control input and its watermark; it uses trusted proprioceptive feedback to generate target poses that bring the physical end-effector velocity to zero in a damped fashion.

\subsection{Replay-Attack Model and MCUSUM Detection and Belief}

\paragraph{Replay-Attack Model}
Replay attacks seek to remain statistically consistent with nominal operation by replacing the current visual stream with previously recorded legitimate observations while the physical robot evolves under potentially manipulated control inputs~\cite{aftabi2025dynamark}. We assume that the adversary has access to the sensing channels and can observe and replace the RGB-D observations received by the controller and detector. The adversary follows a two-stage strategy: first, it records a sufficiently long benign trajectory $x_t$ during closed-loop operation without active intervention; subsequently, it replays the recorded observations. Thus, during replay, the controller and detector receive the spoofed observation $x_t'$, whereas the physical robot receives the adversarial control input $F(x_t') = u_t^{a}$ that is predicted by the control policy on spoofed observations. In the considered threat model, it is assumed that the adversary cannot directly observe or measure the secret dynamic-watermarking signal injected into the control channel, meaning their replayed observations fail to account for the real-time physical excitation.

\paragraph{Multi-Horizon Latent Residual and MCUSUM Detector}
At each detector update, the latent dynamics network predicts an $H$-step trajectory conditioned on the current latent state and applied watermarked control. We consider disjoint prediction windows with stride $H$.
Define the predicted and observed latent trajectories for window $j$ as $\widetilde{\mathbf{z}}_{t_j}:=\mathrm{vec}\left(\tilde z_{t_j-H+1|t_j-H}, \dots, \tilde z_{t_j|t_j-H}\right)\in\mathbb{R}^{Hd_z}$ and $\mathbf{z}_{t_j}:=\mathrm{vec}\left(z_{t_j-H+1}, \dots, z_{t_j}\right)\in\mathbb{R}^{Hd_z}$, respectively. Here, $t_j$ denotes the physical sample at which window $j$ is completed, such that $t_j-t_{j-1}=H$. The corresponding multi-horizon latent residual is
\begin{equation}\label{eq:latent_residual}
    \mathbf{r}_{j} := \mathbf{z}_{t_j} - \widetilde{\mathbf{z}}_{t_j}\in\mathbb{R}^{Hd_z}.
\end{equation}

The nominal residual distribution is calibrated from nominal trajectories. Let $\mu_0$ and $\Sigma_0$ denote the calibrated residual mean and covariance, respectively, and let $\Sigma_0 = \Gamma \Gamma^\top$ be its Cholesky factorization. The standardized residual supplied to the detector is
\begin{equation}\label{eq:whitened_residual}
    \pmb{\varepsilon}_j := \Gamma^{-1}\left(\mathbf{r}_j - \mu_0\right).
\end{equation}
Under nominal operation, $\pmb{\varepsilon}_j\sim\mathcal{N}(\mathbf{0},I_{Hd_z}).$

To accumulate persistent deviations from this nominal model, we employ a multivariate cumulative-sum (MCUSUM) detector. Let $\mathbf{s}_{j-1}\in\mathbb{R}^{Hd_z}$ be the MCUSUM state memory and $\mathbf{s}_0=\mathbf{0}$. For reference parameter $q>0$, the recursion is
\begin{align}
    \mathbf{d}_j &= \mathbf{s}_{j-1}+\pmb{\varepsilon}_j,
    \label{eq:mcusum_dt}\\
    \mathbf{s}_j &=
    \begin{cases}
        \mathbf{0}, & \|\mathbf{d}_j\|_2\le q,\\
        \displaystyle
        \left(1-\frac{q}{\|\mathbf{d}_j\|_2}\right) \mathbf{d}_j, & \|\mathbf{d}_j\|_2>q,
    \end{cases}
    \label{eq:mcusum_st}\\
    Y_j &= \|\mathbf{s}_j\|_2, \label{eq:mcusum_score}\\
    I_j &= \mathbf{1}\left\{Y_j\ge k\right\}, \label{eq:mcusum_alarm}
\end{align}
where $\mathbf{d}_j$ is the pre-shrinkage detection evidence, combining the accumulated memory $\mathbf{s}_{j-1}$ with the current standardized residual $\pmb{\varepsilon}_j$. The quantity $Y_j$ is the MCUSUM score, $k>0$ is the alarm threshold, and $I_j\in\{0,1\}$ indicates threshold exceedance at $j^{\text{th}}$ detector update. \

Let $\mathcal{H}_0$ denote nominal operation and $\mathcal{H}_A$ denote an active replay attack. Under $\mathcal{H}_0$, define the nominal average run length (ARL$_0$) as $\operatorname{ARL}_0:=\mathbb{E}\left[\inf\left\{j\geq 1:I_j=1\right\} \big| \mathcal{H}_0\right]$. Accordingly, the threshold $k$ is selected to attain a prescribed target $\operatorname{ARL}_0^\star$. We use $\alpha:=1/\operatorname{ARL}_0^\star$ as the corresponding nominal false-alarm frequency per detector update. Because detector windows are disjoint, the corresponding average time to false alarm is $H \cdot \Delta t \cdot \operatorname{ARL}_0^\star$. For an attack beginning at physical sample time $t_a$, detection occurs at the first post-onset detector update $J_d:=\inf\left\{j:t_j\geq t_a,\ I_j=1\right\},$ and the detection delay is $(t_{J_d}-t_a)\Delta t$. Both $\operatorname{ARL}_0$ and the detection delay are first-passage quantities evaluated using the MCUSUM recursion and therefore account for the temporal dependence induced by its state memory.

\paragraph{Residual Distribution under Attack}
Consider a detector window $j$ that lies entirely after the replay onset, such that $t_j-H\geq t_a$. Under $\mathcal{H}_A$, the detector receives the previously recorded initial latent state $z'_{t_j-H}=E(x'_{t_j-H})$ and the recorded latent trajectory $\mathbf{z}'_j$, where primes identify samples from the recorded source segment aligned with the current detector window. The recorded trajectory was generated under the historical applied watermarked-control input sequence $\bar{u}'_{t_j-H:t_j-1} = u'_{t_j-H:t_j-1} + \phi'_{t_j-H:t_j-1}$. Therefore, by~\eqref{eq:ldn}, it satisfies $\mathbf{z}'_j = L^{(H)} ( z'_{t_j-H}, \bar{u}'_{t_j-H:t_j-1} ) + \mathbf{r}'_j$, where $\mathbf r'_j$ denotes the nominal prediction residual associated with the recorded source window. In contrast, the detector evaluates the replayed initial latent state using the watermarked-control input sequence actually applied during the current detector window, $\bar{u}_{t_j-H:t_j-1} = u_{t_j-H:t_j-1}+ \phi_{t_j-H:t_j-1}$. Hence, the residual observed under $\mathcal{H}_A$ is
\begin{equation}\label{eq:attack_residual}
\mathbf{r}_j^A := \mathbf{z}'_j - L^{(H)} \left(z'_{t_j-H}, \bar{u}_{t_j-H:t_j-1} \right).
\end{equation}

Because the LDN is conditioned on the applied watermarked-control input sequence, its prediction accounts for the modeled latent response to the recorded watermark realization. Therefore, the remaining nominal prediction residuals $\mathbf{r}_j$ are prediction error and are independent of the particular watermark realization. Specifically, conditional on the locally matched operating regime and the watermark covariance sequences selected by the policy, $\mathbf{r}'_j$ follows the calibrated nominal residual distribution and is independent of the current and recorded watermark sequences. This assumption does not imply that $\mathbf{z}'_j$ is independent of the recorded watermark. Rather, it assumes that their modeled dependence is captured by $L^{(H)}$.

\begin{lemma}\label{lemma:1}
Suppose that $L^{(H)}$ is continuously differentiable with respect to its control-sequence input, the replayed source window is selected from a locally matched operating regime such that $u'_{t_j-H:t_j-1}\approx u_{t_j-H:t_j-1}$, and the first-order linearization remainder is negligible. Furthermore, assume that $\mathbf{r}'_j$ is independent of $\phi_{t_j-H:t_j-1}$ and $\phi'_{t_j-H:t_j-1}$. Define $B_j:=\nabla_{\mathbf{u}}L^{(H)}(z,\mathbf{u}) \big|_{z=z'_{t_j-H},\mathbf{u}=\bar u_{t_j-H:t_j-1}} \in\mathbb{R}^{Hd_z\times Hd_u}.$
Then, to first order,
\begin{equation}\label{eq:linearized_attack_residual}
\mathbf{r}_j^A \approx \mathbf{r}'_j + B_j\left( \phi'_{t_j-H:t_j-1} - \phi_{t_j-H:t_j-1}\right),
\end{equation}
and hence $\mathbf{r}_j^A \overset{\mathrm{approx.}}{\sim} \mathcal{N}(\mu_0,\Sigma_{A,j}),$
where $\Sigma_{A,j} = \Sigma_0+ B_j(\Sigma'_j+\Sigma_j)B_j^\top$, $\Sigma_j = \operatorname{blkdiag} \left( \Sigma_{t_j-H},\ldots,\Sigma_{t_j-1} \right),$ and $\Sigma'_j = \operatorname{blkdiag} \left( \Sigma'_{t_j-H},\ldots,\Sigma'_{t_j-1} \right).$ Consequently, $\pmb{\varepsilon}_j^A := \Gamma^{-1}(\mathbf r_j^A-\mu_0)$ satisfies $ \pmb{\varepsilon}_j^A \overset{\mathrm{approx.}}{\sim} \mathcal{N}(\mathbf 0,\Omega_{A,j})$
\begin{equation}\label{eq:attacked_standardized_distribution}
\Omega_{A,j} = I_{Hd_z} + \Gamma^{-1}B_j (\Sigma'_j+\Sigma_j) B_j^\top\Gamma^{-\top}.
\end{equation}
\end{lemma}
\begin{proof}
    Substituting the recorded latent trajectory into \eqref{eq:attack_residual} gives $\mathbf{r}_j^A = \mathbf{r}_j' + L^{(H)}(z_{t_j-H}',\bar{u}_{t_j-H:t_j-1}') -L^{(H)}(z_{t_j-H}',\bar{u}_{t_j-H:t_j-1}).$ A first-order expansion of $L^{(H)}$ with respect to the control-sequence input around $(z_{t_j-H}',\bar{u}_{t_j-H:t_j-1})$ yields
    \begin{align*}
        \mathbf{r}_j^A &\approx \mathbf{r}_j' + B_j(\bar{u}_{t_j-H:t_j-1}'-\bar{u}_{t_j-H:t_j-1}) \notag\\
        &\approx \mathbf{r}_j' + B_j(\phi'_{t_j-H:t_j-1}-\phi_{t_j-H:t_j-1}),
    \end{align*}
    where the last approximation follows from $u'_{t_j-H:t_j-1}\approx u_{t_j-H:t_j-1}$. Independence and Gaussianity give the corresponding block covariance for $\phi'_{t_j-H:t_j-1}-\phi_{t_j-H:t_j-1}$, denoted by $\Sigma_j'+\Sigma_j$. Therefore, $\mathbf{r}_j^A \sim \mathcal{N}\left(\mu_0, \Sigma_0+B_j(\Sigma_j'+\Sigma_j)B_j^\top\right)$.

    Finally, applying the nominal standardization in
    \eqref{eq:whitened_residual} on $\mathbf{r}_j^A $ gives
    \begin{align*}
    \mathbb{E}[\pmb{\varepsilon}_j^A] &= \Gamma^{-1}\left(\mathbb{E}[\mathbf{r}_j^A] - \mu_0\right) \approx\mathbf{0},\\
    \operatorname{Cov}(\pmb{\varepsilon}_j^A)&=\Gamma^{-1}\Sigma_{A,j}\Gamma^{-\top}\\
    &= I_{Hd_z} +\Gamma^{-1} B_j (\Sigma'_j+\Sigma_j) B_j^\top\Gamma^{-\top},
    \end{align*}
    where $\Sigma_0=\Gamma\Gamma^\top$ was used in the final equality. The proof is complete.
\end{proof}



\paragraph{MCUSUM Belief Characterization} \label{subsec:belief_characterization}
We introduce the detection confidence
\begin{equation}\label{eq:confidence_definition}
c_j:=\Pr\left(\mathcal{H}_A\mid\mathbf{d}_{1:j}\right), \qquad
c_0:=\Pr(\mathcal{H}_A),
\end{equation}
as the posterior probability of the replay-attack hypothesis given the MCUSUM evidence observed through detector update $j$. By \eqref{eq:mcusum_st} and \eqref{eq:mcusum_alarm}, $\mathbf{s}_j$ and $I_j$ are deterministic functions of $\mathbf{d}_j$ and therefore provide no additional information once $\mathbf{d}_{1:j}$ is observed. Under $\mathcal{H}_A$, the attacker selects an unknown onset time $\tau$ before the system trajectory is realized. Bayes' rule gives the sequential update~\cite{aftabi2025dynamark}
\begin{equation}\label{eq:ct}
c_j = \frac{c_{j-1}\kappa_{1,j}}{c_{j-1}\kappa_{1,j} +(1-c_{j-1})\kappa_{0,j}},
\end{equation}
where $\kappa_{0,j} := f\left(\mathbf{d}_j \mid \mathbf{d}_{1:j-1},\mathcal{H}_0\right)$ and $\kappa_{1,j} := f\left(\mathbf{d}_j \mid \mathbf{d}_{1:j-1},\mathcal{H}_A\right)$ are the conditional likelihoods of the current detection-evidence vector under the nominal and replay-attack hypotheses, respectively.

Because the replay attack may begin at any physical sample, define $A_j:=\{\tau\le t_j-H\},$ which is the event that detector window $j$ lies entirely after the attack onset. We represent onset uncertainty at window resolution by defining $\mathcal{Q}_\tau(j) := \Pr(A_j\mid\mathcal{H}_A),$ $\widetilde{\mathcal{Q}}_\tau(j) := \Pr \left(A_j \mid \mathbf{d}_{1:j-1},\mathcal{H}_A\right),$ and $\mathcal{Q}_\tau^{+}(j) := \Pr\left(A_j \mid \mathbf{d}_{1:j},\mathcal{H}_A\right)$. For $\mathcal{Q}_\tau(j-1)<1$, define the onset hazard between consecutive fully attacked-window events as
\begin{align}\label{eq:onset_hazard}
h_\tau(j) &:= \Pr\left(A_j \mid A_{j-1}^{c},\mathcal{H}_A\right) \notag\\
&= \frac{\mathcal{Q}_\tau(j)-\mathcal{Q}_\tau(j-1)}{1-\mathcal{Q}_\tau(j-1)}.
\end{align}
Windows satisfying $t_j-H<\tau\le t_j$ straddle the attack onset and are excluded from the subsequent analysis, which applies only to fully nominal or fully attacked windows.

\begin{theorem}\label{theorem:mcusum_belief}
    Suppose the standardized-residual distributions in Lemma~\ref{lemma:1} hold, and conditional on the applicable regime, the current standardized residual is independent of $\mathbf{d}_{1:j-1}$. In addition, assume that the onset time is fixed by the attacker before the system trajectory is realized, so that
    \begin{equation} \label{eq:onset_independence}
    \Pr\left(A_j \mid A_{j-1}^{c}, \mathbf{d}_{1:j-1}, \mathcal{H}_A\right) = h_\tau(j).
    \end{equation}
    Define
    \begin{align*}
    f_{0,j} &:= \phi_{Hd_z}\!\left(\pmb{\varepsilon}_j; \mathbf{0},I_{Hd_z} \right) = \phi_{Hd_z}\left( \mathbf{d}_j; \mathbf{s}_{j-1},I_{Hd_z} \right), \\
    f_{A,j} &:= \phi_{Hd_z}\left(\pmb{\varepsilon}_j; \mathbf{0},\Omega_{A,j}\right) = \phi_{Hd_z} \left(\mathbf{d}_j;\mathbf{s}_{j-1},\Omega_{A,j}\right).
    \end{align*}
    Then, the onset probabilities and conditional likelihoods satisfy
    \begin{align}
    \widetilde{\mathcal{Q}}_\tau(j) &= \mathcal{Q}_\tau^{+}(j-1) + \left[1-\mathcal{Q}_\tau^{+}(j-1)\right]h_\tau(j), \label{eq:onset_prior_update}\\
    \kappa_{0,j} &= f_{0,j}, \label{eq:nominal_likelihood}\\
    \kappa_{1,j} &= \left[1-\widetilde{\mathcal{Q}}_\tau(j) \right]f_{0,j} + \widetilde{\mathcal{Q}}_\tau(j) f_{A,j}, \label{eq:attack_likelihood}\\
    \mathcal{Q}_\tau^{+}(j) &= \frac{\widetilde{\mathcal{Q}}_\tau(j) f_{A,j}}{\kappa_{1,j}}. \label{eq:onset_posterior_update}
    \end{align}
\end{theorem}

\begin{proof}
    Given $\mathbf{d}_{1:j-1}$, the previous MCUSUM state $\mathbf{s}_{j-1}$ is known. Moreover, $\pmb{\varepsilon}_j = \mathbf{d}_j - \mathbf{s}_{j-1}.$ For fixed $\mathbf{s}_{j-1}$, this transformation is a translation with unit Jacobian. Therefore,
    \begin{align*}
    \kappa_{0,j} &= f\left(\mathbf{d}_j \mid \mathbf{d}_{1:j-1},\mathcal{H}_0\right)\\
    &=f_{\pmb{\varepsilon}} \left(\pmb{\varepsilon}_j \mid \mathbf{d}_{1:j-1},\mathcal{H}_0 \right)\\
    &= \phi_{Hd_z}\left(\pmb{\varepsilon}_j; \mathbf{0},I_{Hd_z}\right) = f_{0,j},
    \end{align*}
    where the final equality follows from the conditional-independence
    assumption. This proves \eqref{eq:nominal_likelihood}. Similarly,
    Lemma~\ref{lemma:1} gives
    \begin{equation}
    f\left(\mathbf{d}_j \mid \mathbf{d}_{1:j-1}, A_j,\mathcal{H}_A \right)= \phi_{Hd_z}\left(\pmb{\varepsilon}_j;
    \mathbf{0},\Omega_{A,j} \right)=f_{A,j}.
    \label{eq:active_window_likelihood}
    \end{equation}
    For a pre-onset window, the same argument under the nominal residual model gives
    \begin{equation}
    f\left(\mathbf{d}_j \mid \mathbf{d}_{1:j-1}, A_j^c,\mathcal{H}_A \right) = f_{0,j}.
    \label{eq:pre_onset_window_likelihood}
    \end{equation}
    Because $A_{j-1}\subseteq A_j$, the event $ A_j$ is the disjoint union $ A_j= A_{j-1}\cup\left( A_j\cap A_{j-1}^c\right).$ Consequently,
    \begin{align*}
    \widetilde{\mathcal{Q}}_\tau(j) =& \Pr\left(A_j \mid \mathbf{d}_{1:j-1},\mathcal{H}_A \right)\\
    =& \Pr\left(A_{j-1} \mid \mathbf{d}_{1:j-1},\mathcal{H}_A\right) \\
    &+ \Pr\left(A_j \cap A_{j-1}^{c} \mid \mathbf{d}_{1:j-1},\mathcal{H}_A\right) \\
    =& \mathcal{Q}_\tau^{+}(j-1) \\
    &+\left[1-\mathcal{Q}_\tau^{+}(j-1)\right] \Pr\left(A_j \mid A_{j-1}^{c}, \mathbf{d}_{1:j-1}, \mathcal{H}_A \right)\\
    =& \mathcal{Q}_\tau^{+}(j-1) + \left[1-\mathcal{Q}_\tau^{+}(j-1)\right] h_\tau(j),
    \end{align*}
    which proves \eqref{eq:onset_prior_update}.
    Conditioning the attack-hypothesis likelihood on $A_j$ gives
    \begin{align*}
    \kappa_{1,j} =& f\left(\mathbf{d}_j \mid \mathbf{d}_{1:j-1},\mathcal{H}_A \right)\\
    =& f\left(\mathbf{d}_j \mid \mathbf{d}_{1:j-1}, A_j^c,\mathcal{H}_A \right) \Pr\left(A_j^c \mid \mathbf{d}_{1:j-1},\mathcal{H}_A \right)\\
    &+ f\left(\mathbf{d}_j \mid \mathbf{d}_{1:j-1}, A_j,\mathcal{H}_A\right) \Pr\left(A_j \mid \mathbf{d}_{1:j-1}, \mathcal{H}_A \right)\\
    =& \left[1 - \widetilde{\mathcal{Q}}_\tau(j)\right] f_{0,j} + \widetilde{\mathcal{Q}}_\tau(j) f_{A,j},
    \end{align*}
    where \eqref{eq:active_window_likelihood} and \eqref{eq:pre_onset_window_likelihood} were used. This proves \eqref{eq:attack_likelihood}. Finally, Bayes' rule within $\mathcal H_A$ gives
\begin{align*}
\mathcal{Q}_\tau^{+}(j) &= \Pr\left(A_j \mid \mathbf{d}_{1:j},\mathcal{H}_A \right)\\
&= \frac{f\left(\mathbf{d}_j \mid \mathbf{d}_{1:j-1}, A_j,\mathcal{H}_A \right) \Pr\left(A_j \mid \mathbf{d}_{1:j-1},\mathcal{H}_A \right)}{f\left(\mathbf{d}_j \mid \mathbf{d}_{1:j-1},\mathcal{H}_A\right)}\\
&= \frac{\widetilde{\mathcal{Q}}_\tau(j) f_{A,j}}{\kappa_{1,j}},
\end{align*}
which proves \eqref{eq:onset_posterior_update}.
\end{proof}






\subsection{Adaptive Dynamic Watermarking via Reinforcement Learning}\label{subsec:dynamark_rl}
DynaMark adapts the distribution of the private watermark without
replacing the nominal image-based controller $F$. We formulate its
adaptive covariance selection as a Markov Decision Process
(MDP)~\cite{Puterman528623,sutton2018reinforcement}. At time $t$, the
DynaMark information state is
\begin{equation}\label{eq:dm_state}
    s_t
    :=
    (z_t,u_t,\Sigma_{t-1},c_t)\in\mathcal{S},
\end{equation}
where $z_t$ and $u_t$ are the current latent state and control input respectively, $\Sigma_{t-1}$ is the predicted DWM covariance at the previous step and  $c_t\in[0,1]$ is the detection confidence from the belief update in~\eqref{eq:ct}. The DynaMark policy is the map
\begin{equation}\label{eq:dm_interface}
    DM:\mathcal{S}\to\mathbb{S}_+^{d_u},
    \qquad
    \Sigma_t=DM(s_t).
\end{equation}
The selected covariance determines
$\phi_t\sim\mathcal{N}(0,\Sigma_t)$, which is the watermark signal added to the nominal control input through~\eqref{eq:u_phi}. The resulting observation is encoded and compared with the LDN prediction, producing new detector evidence
and the subsequent confidence supplied back to $DM$. 

\paragraph{Action Space}
The action is the positive semidefinite watermark covariance
$\Sigma_t$. Its magnitude is bounded to limit the excitation applied
to the nominal control:
\begin{equation}\label{eq:action_space}
    \mathcal{A}(s_t)
    =
    \left\{
        \Sigma \in \mathbb{S}_+^{d_u} :
        \|\Sigma\|_F \le \Sigma_{\max}
    \right\}.
\end{equation}
Here, $\Sigma_{\max}>0$ is chosen according to the desired excitation level. Task-specific safety is incorporated through the reward rather than as an additional action-space constraint.

\paragraph{Reward Function}
The reward balances watermark expenditure, deviation from nominal
behavior, detector confidence, and robot--human
separation. Let
$\tilde z_{t+1}^{\mathrm{nom}}:=L(z_t,u_t)$ denote the predicted
successor under the nominal control without watermarking, and let
$\omega_1,\omega_2,\omega_3>0$ be design weights.

The base watermarking reward is
\begin{align}\label{eq:base_wm_reward}
    r_t^{\mathrm{wm}}
    =&-\omega_1\|\phi_t\|_1
      -\omega_2\|\tilde z_{t+1}^{\mathrm{nom}}-z_{t+1}\|_2
      \notag\\
     &+\omega_3|0.5-c_{t+1}|.
\end{align}
The first term penalizes watermark magnitude, the second penalizes deviation from the predicted unwatermarked evolution, and the third rewards detector confidence away from $0.5$. DWM training includes both clean and replay-attacked episodes, exposing the policy to both operating conditions. The intended response is low attack confidence during clean operation and high confidence after replay onset. The confidence term encourages decisive beliefs, while the other terms limit the motion and excitation costs of watermarking. The reward uses the subsequent confidence $c_{t+1}$ to evaluate the effect of the selected covariance.

To incorporate safety into policy optimization, we define a safety metric ($\beta_t$) specific to our use case (human-robot interaction), observed after applying $\bar u_t$
\begin{equation}\label{eq:safety_metric}
    \beta_{t+1}
    =
    \|P_{r,t+1}-P_{h,t+1}\|_2,
\end{equation}
where $P_{r,t+1}$ and $P_{h,t+1}$ denote the points on the robot end-effector
and human-arm that are closest to one another, respectively.
For a minimum desired separation $\beta_\text{min}>0$, define the
nonnegative safety cost
\begin{equation}\label{eq:safety_cost}
    r_t^{\mathrm{safe}}
    =
    [\beta_\text{min}-\beta_{t+1}]_+,
    \qquad [a]_+:=\max(0,a).
\end{equation}
Here, $a\in\mathbb{R}$ is a scalar argument of the positive-part
operator $[a]_+$. In~\eqref{eq:safety_cost}, $a=\beta_\text{min}-\beta_{t+1}$
represents the separation shortfall, so the safety cost is positive
only when the observed separation falls below $\beta_\text{min}$.
The reward used during policy optimization is then
\begin{equation}\label{eq:lagrangian_reward}
    r_t^\lambda(s_t,\Sigma_t)
    =
    r_t^{\mathrm{wm}}
    -
    \lambda \cdot r_t^{\mathrm{safe}},
    \qquad \lambda\ge0.
\end{equation}
Thus, the covariance bound is enforced through the action space, while separation violations are penalized during training. The multiplier $\lambda$ sets the weight of the safety cost. As training progresses and the policy learns to reduce separation violations, $\lambda$ is reduced while remaining nonnegative.

\begin{figure}[t]
    \centering
    \vspace*{6pt}
    \includegraphics[width=0.95\columnwidth]{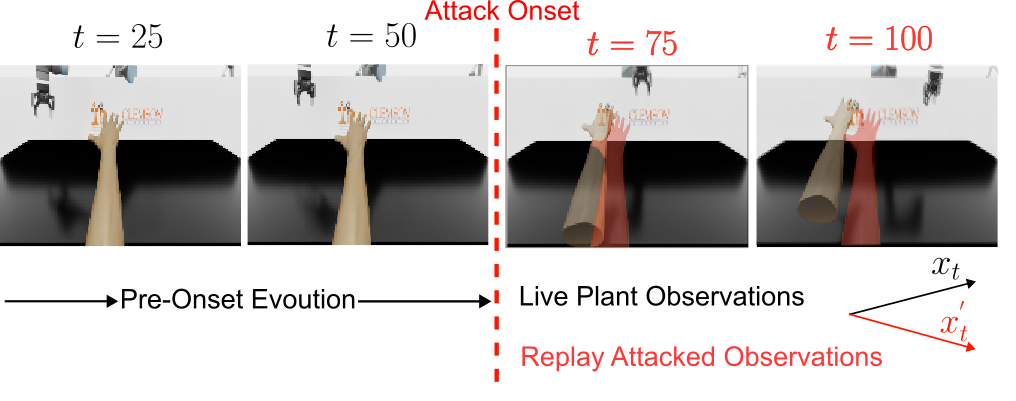}
    \caption{Live plant observations and replayed controller observations}
    \label{fig:replay_vs_true_sim}
\end{figure}

\section{Experiments}
\label{sec:experiments}

\subsection{Human-Robot Interactive Task Environment}
We evaluate a vision-guided robotic manipulator in Isaac Sim while it reaches a target and avoids a moving human arm. The controller $F$ receives RGB-D observation $x_t\in\mathbb{R}^{120\times160\times4}$, joint positions and velocities, end-effector position, and goal position form $x_t^{\mathrm{prop}}\in\mathbb{R}^{18}$. The nominal controller $F$ outputs a three-dimensional action $u_t\in[-1,1]^3$. Figure~\ref{fig:replay_vs_true_sim} illustrates the live-versus-replay attacked visual discrepancy before and after the recorded replay onset. The policy controller runs at 30Hz, thus $\Delta t \approx 0.033s$. 

\subsection{Latent Encoder and DWM Policy Training Setup}
We train a V-JEPA-inspired~\cite{assran2025vjepa2} encoder and action-conditioned LDN on clean nominal and watermarked RGB-D trajectories, excluding replay data with latent dimension $d_z=64$ and LDN prediction horizon $H=4$. 
With $F$ frozen, we train the DWM policy using PPO~\cite{schulman2017proximal} on both clean and replay-attacked episodes. The policy predicts a watermarking covariance $\Sigma_t=\operatorname{diag}(\sigma_t^2)$, such that $0\leq\sigma_{t,i}\leq0.3$ according to \eqref{eq:action_space}. The objective balances detector confidence, excitation and motion costs, and the Lagrangian separation penalty. Latent residual windows $(\mathbf{r}_j)$ are obtained from encoded RGB-D images (replay vs predicted), watermark covariances are selected by the trained PPO policy, and attack beliefs are updated using the likelihoods defined for~\eqref{eq:ct}.
\subsection{Replay Protocol and Detector Calibration}
In paired-fork rollouts, the target and source start from matched states; at $t_a$, the target receives the source RGB-D stream while its physical trajectory continues. Disjoint $H$-step windows yield residuals $\mathbf r_j\in\mathbb{R}^{Hd_z}$ from~\eqref{eq:latent_residual}. Separate clean episodes estimate $(\mu_0,\Sigma_0)$ and calibrate $k$ against a prescribed nominal average run length $\mathrm{ARL}_0$. At each update, the LDN uses the current private control sequence and its Jacobian to form the replay covariance; the detector uses $q=0.5$, $c_0=0.01$, and a geometric onset hazard of $10^{-3}$ per window. Continuous likelihoods update $c_j$ independently of the MCUSUM alarm. Calibration and evaluation episodes remain disjoint, and live ground truth is withheld from the detector.

\section{Results}
\label{sec:results}

\begin{table}[t]
    \centering
    \caption{Replay Detection and Safety Test
    (30 episodes).}
    \label{tab:ur5_results}
    \small
    \setlength{\tabcolsep}{3pt}
    \begin{tabular}{lcc}
        \hline
        \multicolumn{3}{c}{\textbf{Replay Detection}} \\
        \hline
        Metric & No WM & Dynamic WM \\
        \hline
        Clean Episode False Alarms & 0/30 & 6/30 \\
        Belief based Attack Detection & 0/30 & 29/30 \\
        Median detection delay (s) & --- & 1.23 \\
        Early Belief Crossing & 0/30 & 1/30 \\
        \hline
        \multicolumn{3}{c}{\textbf{Safety under Replay with Watermarking}} \\
        \hline
        Metric & No Simplex & Simplex \\
        \hline
        Episodes below 3~cm & 20/30 & 0/30 \\
        Minimum clearance (m) & 0.023 & 0.122 \\
        Final clearance (m) & 0.074 & 0.210 \\
        Final mean speed (m/s) & 0.140 & 0.003 \\
        \hline
    \end{tabular}
\end{table}

\subsection{Replay Detection and State Divergence}
Replay separates the received observations from the physical plant, as illustrated in Figure.~\ref{fig:replay_vs_true_sim}. Private watermarking tests whether that feedback contains the response to the current excitation. In Table~\ref{tab:ur5_results}, 29 of 30 watermarked replay episodes crossed the belief threshold after onset without an earlier crossing, with median delay 1.23~s. The remaining episode crossed both before and after onset. Six clean watermarked episodes also crossed the belief threshold, indicating a specificity limitation rather than alarms rejected by belief accumulation. Without watermarking, none of the 30 replay episodes crossed the belief threshold, supporting the role of private excitation in this evaluation.

\begin{figure}[t]
    \centering
    \includegraphics[width=\columnwidth]{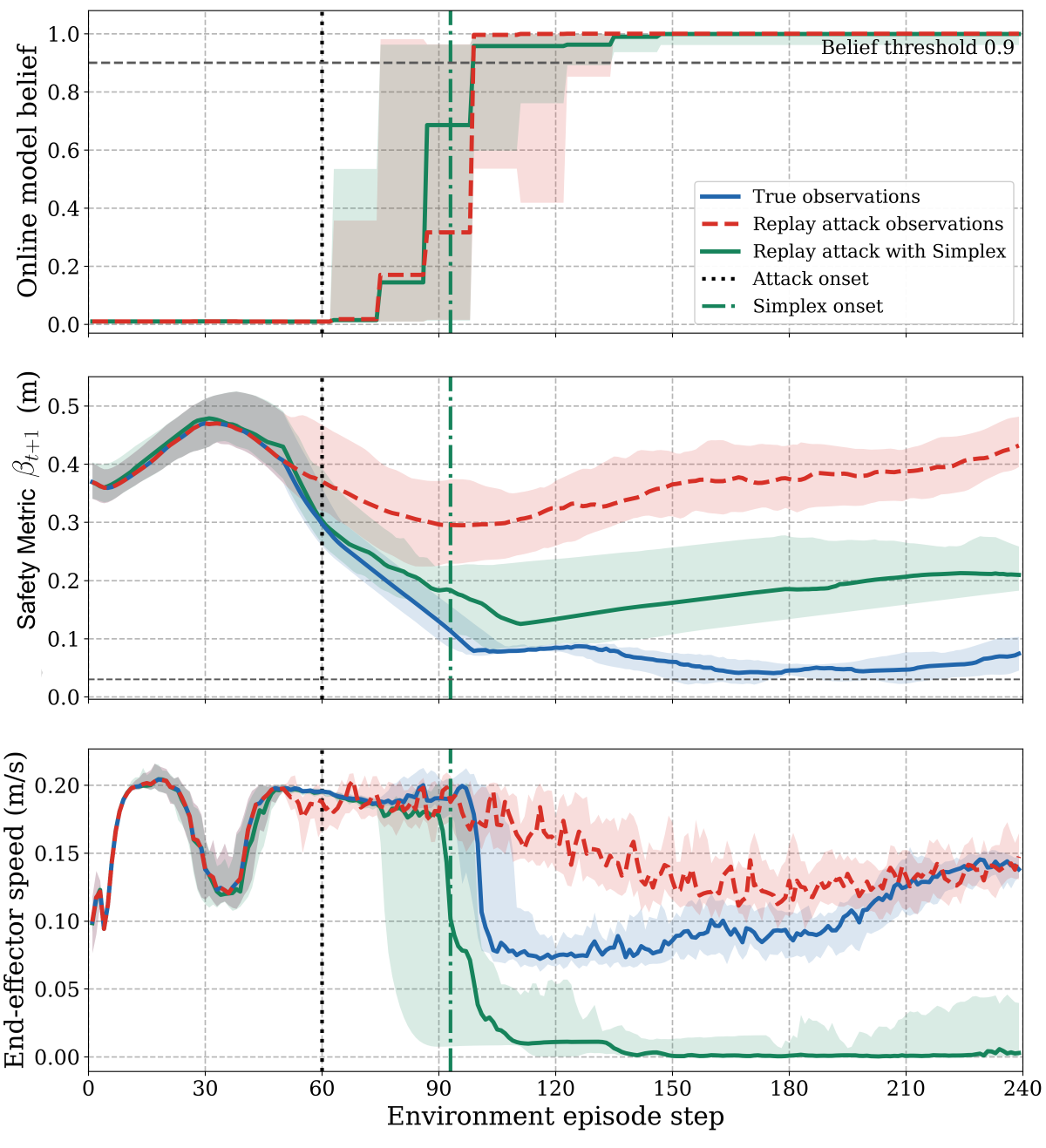}
    \caption{Average \textit{(Top)} Attack Belief, \textit{(Middle)} Min Robot-Human separation (m), and \textit{(Bottom)} End-effector speed (m/s) across 30 episodes. Replay begins at step 60. Blue denotes live observations, red denotes the replayed source observations received by the policy, and green denotes the live trajectory with Simplex stopping.}
    \label{fig:belief}
\end{figure}

\subsection{Safety Benefits of Simplex Switching}
Detection alone does not prevent the robot from following misleading feedback. A separate test therefore evaluates Simplex switching~\eqref{eq:supervisory_command} under replay with watermarking. From Figure~\ref{fig:belief}, in the belief panel \textit{(Top)}, the red curve shows confidence from the replayed observation stream and the green curve shows confidence in the replay experiment with Simplex. In the Safety metric \textit{(Middle)} and end effector speed plots \textit{(Bottom)}, blue describes the physical robot without Simplex, red describes the replayed source trajectory, and green describes the physical robot with Simplex. It is clear that the policy visible, replay attacked observations show the safety metric to be increasing, whereas in reality the true safety metric actually decreases caused by the attack. With Simplex active, the live robot maintains greater separation from the human arm and its end-effector speed approaches zero. The safety comparisons in Table~\ref{tab:ur5_results} use the live physical trajectories, rather than the replayed source trajectory. 

Simplex switches to stopping when attack confidence exceeds $0.9$ for the required persistence period. As shown in Figure.~\ref{fig:belief} and Table~\ref{tab:ur5_results}, 20 of 30 episodes without Simplex entered the 3~cm proximity zone, compared with none with Simplex. The reported minimum clearance increased from 0.023 to 0.122~m, and final speed decreased from 0.140 to 0.003~m/s. These results show that the fallback reduces continued physical motion under misleading visual feedback. Of 30 switches, 28 followed attack onset and two occurred early. Watermarking supplies attack evidence, and Simplex uses that evidence to initiate a protective stop. A limitation remains that false attack detections can still interrupt the task and reduce performance.

\section{Conclusion}
This paper combines private dynamic watermarking, action-conditioned latent dynamics, Bayesian attack confidence, and Simplex stopping for vision-based control. The detector compares latent predictions conditioned on the applied watermarked commands with representations encoded from the received RGB-D images. A trained PPO policy selects the watermark covariance, while a Lagrangian separation penalty discourages watermark choices that bring the robot too close to the human arm. The analysis derives an approximate replay-residual covariance and a sequential belief update under the stated modeling assumptions.

Reach-avoid experiments evaluate this visual detector and trained watermark policy. Watermarking produced post-onset belief crossings without earlier crossings in 29 of 30 replay episodes, with a median delay of 1.23~s. Simplex stopping reduced episodes entering the 3~cm proximity zone from 20 of 30 to none in the separate safety test. Six clean episodes crossed the belief threshold, and two switches in the safety test occurred before attack onset, showing that unnecessary intervention remains a limitation. These results demonstrate replay detection and reduced motion under misleading feedback within the evaluated task; the training penalty and stopping response do not establish a formal safety guarantee.

Future work will use neural network verification tools to develop formal safety guarantees, extend the framework to physical robot experiments with sensing and actuation delays, and improve latent dynamics prediction to reduce false alarms. Further evaluation will also consider continued human motion after the robot stops.

\section*{Acknowledgments}
The authors used OpenAI's ChatGPT~\cite{openai_chatgpt}
throughout the manuscript to assist with writing, proofreading,
and minor language edits for grammar, clarity, and readability.
All images were devised by the authors, who also conducted
the experiments and generated the plots. The authors reviewed
and approved the final text and take full responsibility for
the manuscript's content and accuracy.

\bibliography{ref}

@inproceedings{korshunov2024replay,
  author    = {Korshunov, Pavel and George, Anjith and
               {\"O}zbulak, G{\"o}khan and Marcel, S{\'e}bastien},
  title     = {Vulnerability of Face Age Verification to Replay Attacks},
  booktitle = {2024 IEEE International Conference on Acoustics,
               Speech and Signal Processing (ICASSP)},
  year      = {2024},
  pages     = {4830--4834},
  doi       = {10.1109/ICASSP48485.2024.10447255}
}

@article{khan2025spatiotemporal,
  author  = {Khan, Shujaat and Siddique, Taha Hasan Masood and
             Ibrahim, Muhammad Sohail and Siddiqui, Abdul Jabbar and
             Huang, Kejie},
  title   = {Spatio-temporal deep learning for improved face
             presentation attack detection},
  journal = {Knowledge-Based Systems},
  volume  = {311},
  pages   = {113059},
  year    = {2025},
  doi     = {10.1016/j.knosys.2025.113059}
}

@misc{openai_chatgpt,
  author       = {{OpenAI}},
  title        = {{ChatGPT}},
  howpublished = {Online. Available: https://chatgpt.com},
  note         = {Accessed: September 29, 2026}
}

@article{schulman2017proximal,
  author        = {Schulman, John and Wolski, Filip and Dhariwal, Prafulla and Radford, Alec and Klimov, Oleg},
  title         = {Proximal Policy Optimization Algorithms},
  journal       = {arXiv preprint arXiv:1707.06347},
  year          = {2017},
  eprint        = {1707.06347},
  archivePrefix = {arXiv},
  primaryClass  = {cs.LG},
  doi           = {10.48550/arXiv.1707.06347}
}

@book{Puterman528623,
author = {Puterman, Martin L.},
title = {Markov Decision Processes: Discrete Stochastic Dynamic Programming},
year = {1994},
isbn = {0471619779},
publisher = {John Wiley \& Sons, Inc.},
address = {USA},
edition = {1st}
}

@book{sutton2018reinforcement,
author = {Sutton, Richard S. and Barto, Andrew G.},
title = {Reinforcement Learning: An Introduction},
year = {2018},
isbn = {0262039249},
publisher = {A Bradford Book},
address = {Cambridge, MA, USA}
}

@article{satchidanandan2017dynamic,
author = {Satchidanandan, Bharadwaj and Kumar, P. R.},
title = {Dynamic Watermarking: Active Defense of Networked Cyber-Physical Systems},
journal = {Proceedings of the IEEE},
volume = {105},
number = {2},
pages = {219--240},
year = {2017}
}

@article{naha2023sequential,
author = {Naha, Abhishek and Teixeira, Andre M. H. and Ahl\'en, Anders and Dey, Subhrakanti},
title = {Sequential Detection of Replay Attacks},
journal = {IEEE Transactions on Automatic Control},
volume = {68},
number = {3},
pages = {1941--1948},
year = {2023},
doi = {10.1109/TAC.2022.3174004}
}

@inproceedings{mo2009secure,
author = {Mo, Yilin and Sinopoli, Bruno},
title = {Secure Control Against Replay Attacks},
booktitle = {2009 47th Annual Allerton Conference on Communication, Control, and Computing},
pages = {911--918},
year = {2009},
doi = {10.1109/ALLERTON.2009.5394956}
}

@article{porter2021generalized,
author = {Porter, Matthew and Hespanhol, Pedro and Aswani, Anil and Johnson-Roberson, Matthew and Vasudevan, Ram},
title = {Detecting Generalized Replay Attacks via Time-Varying Dynamic Watermarking},
journal = {IEEE Transactions on Automatic Control},
volume = {66},
number = {8},
pages = {3502--3517},
year = {2021}
}

@ARTICLE{aftabi2025dynamark,
  author={Aftabi, Navid and Hanchate, Abhishek and Bukkapatnam, Satish T. S. and Li, Dan},
  journal={IEEE Transactions on Automation Science and Engineering}, 
  title={DynaMark: A Reinforcement Learning Framework for Dynamic Watermarking in Industrial Machine Tool Controllers}, 
  year={2026},
  volume={23},
  number={},
  pages={13658-13676},
  doi={10.1109/TASE.2026.3715642}}

@inproceedings{wu2023daydreamer,
author = {Wu, Philipp and Escontrela, Alejandro and Hafner, Danijar and Abbeel, Pieter and Goldberg, Ken},
title = {DayDreamer: World Models for Physical Robot Learning},
booktitle = {Proceedings of the Conference on Robot Learning},
series = {Proceedings of Machine Learning Research},
volume = {205},
pages = {2226--2240},
year = {2023},
publisher = {PMLR}
}

@InProceedings{hafner2019planet,
  author = {Hafner, Danijar and Lillicrap, Timothy and Fischer, Ian and Villegas, Ruben and Ha, David and Lee, Honglak and Davidson, James},
  title = {Learning Latent Dynamics for Planning from Pixels},
  booktitle = {Proceedings of the 36th International Conference on Machine Learning},
  pages = {2555--2565},
  year = {2019},
  volume = {97},
  series = {Proceedings of Machine Learning Research},
  publisher = {PMLR}
}

@InProceedings{hafner2020dreamer,
  author = {Hafner, Danijar and Lillicrap, Timothy and Ba, Jimmy and Norouzi, Mohammad},
  title = {Dream to Control: Learning Behaviors by Latent Imagination},
  booktitle = {International Conference on Learning Representations},
  year = {2020}
}

@InProceedings{hansen2022tdmpc,
  author = {Hansen, Nicklas A. and Su, Hao and Wang, Xiaolong},
  title = {Temporal Difference Learning for Model Predictive Control},
  booktitle = {Proceedings of the 39th International Conference on Machine Learning},
  pages = {8387--8406},
  year = {2022},
  volume = {162},
  series = {Proceedings of Machine Learning Research},
  publisher = {PMLR}
}

@misc{assran2025vjepa2,
author = {Assran, Mahmoud and others},
title = {{V-JEPA 2}: Self-Supervised Video Models Enable Understanding, Prediction and Planning},
year = {2025},
eprint = {2506.09985},
archivePrefix = {arXiv},
primaryClass = {cs.AI},
url = {https://arxiv.org/abs/2506.09985}
}

@inproceedings{parameshwaran2025sevin,
author = {Parameshwaran, Aditya and Wang, Yue},
title = {Scalable and Interpretable Verification of Image-Based Neural Network Controllers for Autonomous Vehicles},
booktitle = {Proceedings of the ACM/IEEE 16th International Conference on Cyber-Physical Systems (ICCPS)},
year = {2025},
doi = {10.1145/3716550.3722037},
url = {https://doi.org/10.1145/3716550.3722037}
}
\bibliographystyle{IEEEtran}

\end{document}